\documentclass[11pt,letterpaper]{article}
\usepackage{emnlp2016}
\usepackage{times}
\usepackage{amsmath}
\usepackage{graphicx}
\usepackage{fancyhdr}
\usepackage{adjustbox}
\usepackage{multirow}
\usepackage{float}
\usepackage{textcomp}
\usepackage{algorithm,algorithmic}
\usepackage{amssymb,amsthm}
\usepackage{bbm}

\emnlpfinalcopy

\theoremstyle{definition}
\newtheorem{defn}{Definition}

\DeclareMathOperator*{\argmax}{arg\,max} 

\makeatletter
\newcommand{\ALOOP}[1]{\ALC@it\algorithmicloop\ #1%
	\begin{ALC@loop}}
	\newcommand{\ENDALOOP}{\end{ALC@loop}\ALC@it\algorithmicendloop}

\newcommand{\algorithmicbreak}{\textbf{break}}
\newcommand{\BREAK}{\STATE \algorithmicbreak}
\makeatother

\title{Supervised Opinion Aspect Extraction by Exploiting Past Extraction Results}

\author{Lei Shu$^{1}$, Bing Liu$^1$, Hu Xu$^1$, Annice Kim$^2$\\
	$^1$Department of Computer Science, University of Illinois at Chicago, USA\\
	$^2$Center for Health Policy Science and Tobacco Research, RTI International, USA\\
	$^{1}$\{lshu3, liub, hxu48\}@uic.edu, $^2$akim@rti.org\\
}

\begin{document}

\maketitle
\begin{abstract}
One of the key tasks of sentiment analysis of product reviews is to extract product aspects or features that users have expressed opinions on. In this work, we focus on using supervised sequence labeling as the base approach to performing the task. Although several extraction methods using sequence labeling methods such as Conditional Random Fields (CRF) and Hidden Markov Models (HMM) have been proposed, we show that this supervised approach can be significantly improved by exploiting the idea of concept sharing across multiple domains. For example, ``screen'' is an aspect in iPhone, but not only iPhone has a screen, many electronic devices have screens too. When ``screen'' appears in a review of a new domain (or product), it is likely to be an aspect too. Knowing this information enables us to do much better extraction in the new domain. This paper proposes a novel extraction method exploiting this idea in the context of supervised sequence labeling. Experimental results show that it produces markedly better results than without using the past information.
\end{abstract}

\section{Introduction}
\label{sec:introduction}
Aspect extraction is a fundamental task of opinion mining or sentiment analysis \cite{HuL2004}. It aims to extract opinion targets from opinion text. For example, from ``\emph{This phone has a good screen},'' it aims to extract ``screen.'' In product reviews, aspects are product attributes or features. They are needed in many sentiment analysis applications.

Aspect extraction has been studied by many researchers. There are both \emph{supervised} and \emph{unsupervised} approaches. We will discuss existing methods in these two approaches and compare them with the proposed technique in the related work section. The proposed technique uses the popular supervised sequence labeling method Conditional Random Fields (CRF) \cite{Lafferty2001conditional} as its base algorithm  \cite{Jakob2010,Choi2010,Mitchell-EtAl:2013:EMNLP}. We will show that the results of CRF can be significantly improved by leveraging some prior knowledge mined automatically (without any user involvement) from a large amount of online reviews of many products, which we also call domains. Such reviews are readily available and can be easily crawled from the Web, e.g., from Amazon.com's review pages. The improvement is possible due to the important observation that although every product (domain) is different, there is a fair amount of aspects overlapping across domains or products~\cite{ChenLiu2014ICML}. For example, every product review domain probably has the aspect \textit{price}, reviews of most electronic products share the aspect of \textit{battery life} and reviews of some products share the aspect of \textit{screen}. This paper exploits such sharing to help CRF produce much better extraction results. 


Since the proposed method will make use of the past extraction results as prior knowledge to help new or future extraction, it has to make sure that the knowledge is reliable. It is well-known that no statistical learning method can guarantee perfect results (as we will see later in the experiment section, CRF's extraction results are far from perfect). However, if we can find a set of shared aspects that have been extracted from multiple past domains, these aspects, which we call {\em reliable aspects}, are more likely to be correct. They can serve as the prior knowledge to help CRF extract from a new domain more effectively. For example, we have product reviews from three domains. After running a CRF-based extractor, a set of aspects is extracted from each domain reviews, which is listed below. Note that only four aspects are listed for each domain for illustration purposes.

~~~~Camera Domain: \textit{price}, \textit{my wife}, \textit{battery life}, \textit{picture}

~~~~Cellphone: \textit{picture}, \textit{husband}, \textit{battery life}, \textit{expensive}

~~~~Washer: \textit{price}, \textit{water}, \textit{customer}, \textit{shoes}

\noindent
Clearly, some of the aspects are clearly incorrect, e.g., {\em my wife}, {\em husband}, and {\em customer} as they are not features of these products. However, if we focus on those aspects that appear at least in two domains, we can find the following set:

~~~~\{\textit{price}, \textit{battery life},  \textit{picture}\}.

\noindent
This list of words is used as the past knowledge and given to CRF, which will leverage on it to perform better extraction in the new review domain. 


The proposed approach has two phases:

\begin{enumerate}
	\item {\em Model building phase}: Given a labeled training review dataset $D_t$, it builds a CRF model $M$. 
	\item {\em Extraction phase}:  At any point in time, $M$ has been applied to extract from $n$ past domains of reviews $D=\{D_1,\dots,D_n\}$, which produced the corresponding sets of aspects $S=\{A_1,\dots, A_n\}$. A set of frequent aspects $K$ (the past knowledge) will be discovered from $S$. When faced with a new domain of reviews $D_{n+1}$, the algorithm first finds a set $K$ of reliable aspects from the aspect store $S$. $K$ is then used to help $M$ perform better extraction from $D_{n+1}$. The resulting set of aspects $A_{n+1}$ is added to $S$ for future use. 
	
\end{enumerate}


Due to the ability to leverage the knowledge $K$ gained from the past learning results to help the new domain extraction by the CRF model $M $. We are essentially using the idea of {\em lifelong machine learning}~\cite{ChenLiu2014ICML,ruvolo2013ella,silver2013lifelong,thrun1998lifelong}, which gives the name of the proposed technique, {\em Lifelong-CRF}. Lifelong learning means to retain the knowledge or information learned in the past and leverage it to help future learning and problem solving. Formally, it is defined as follows~\cite{chen2015lifelong}: 




\begin{defn}
	\label{def:lifelong_machine_learning}
	{\em Lifelong machine learning} (or simply lifelong learning) is a continuous learning process where the learner has performed a sequence of $n$ learning tasks, $T_1$, $T_2$, $\dots$, $T_{n}$, called the {\em past tasks}.  When faced with the $(n+1)$th task $T_{n+1}$ with its data $D_{n+1}$, the learner can leverage the {\em prior knowledge} $K$ gained in the past to help learn $T_{n+1}$. After the completion of learning $T_{n+1}$, $K$ is updated with the learned results from $T_{n+1}$.	
\end{defn}


The key challenge of the proposed Lifelong-CRF method is how to leverage $K$ to help $M$ to perform better extraction. This paper proposes a novel method, which does not change the trained model $M$, but uses a set of dependency patterns generated from dependency relations and $K$ as feature values for CRF. As the algorithm extracts in more domains, $K$ also grows and the dependency patterns grow too, which gives the new domain richer feature information to enable the CRF model $M$ to perform better extraction in the new domain data $D_{n+1}$. 

In summary, this paper makes the following contributions:

\begin{enumerate}
	
	\item It proposes a novel idea of exploiting review collections from past domains to learn prior knowledge to guide the CRF model in its sequence labeling process. To the best of our knowledge, this is the first time that lifelong learning is added to CRF. It is also the first time that lifelong learning is applied to supervised aspect extraction.
	
	\item It proposes a novel method to incorporate the prior knowledge in the CRF prediction model for better extraction.  
	
	\item Experimental results show that the proposed Lifelong-CRF outperforms baseline methods markedly.  
	
\end{enumerate}

\section{Related Work}



\label{related work}
As mentioned in the introduction, there are two main approaches to aspect extraction: \emph{supervised} and \emph{unsupervised}. The former is mainly based on CRF \cite{Jakob2010,Choi2010,Mitchell-EtAl:2013:EMNLP}, while the latter is mainly based on topic modeling \cite{MeiLWSZ2007,TitovM2008,LiHuangZhu2010,Brody2010,Wang2010,Moghaddam2011,Mukherjee2012,Lin2009,ZhaoJiang2010,Jo2011,FangHuang2012ACL}, and syntactic rules \cite{ZhuangJZ2006,WangBo2008,WuZHW2009,Zhang2010,QiuLBC2011,poria2014rule,xu2016CER,xu2016pcqa}. There are also frequency-based methods \cite{HuL2004,PopescuNE2005,Zhu2009}, word alignment methods \cite{KangLiu2013IJCAI}, label propagation methods \cite{Zhou-wan-xiao:2013:EMNLP}, and others~\cite{zhao2015creating}.


The technique proposed in this paper is in the context of supervised CRF \cite{Jakob2010,Choi2010,Mitchell-EtAl:2013:EMNLP}, which learns a sequence model to label aspects and non-aspects. Our work aims to improve it by exploiting the idea of lifelong learning. None of the existing supervised extraction methods have made use of this new idea. 




Our work is most closely related to extraction methods that have already employed lifelong learning. However, all the current methods are unsupervised. For example, lifelong topic modeling-based methods in \cite{ChenZhiyuan2014ACL,wang2016mining} have been used for aspect extraction. However, topic models can only find some rough topics and are not effective for finding fine-grained aspects as a topical term does not necessarily mean an aspect. 
Also, topic models only find aspects that are individual words, but many aspects of products are multiple word phrases, e.g., {\em batter life} and {\em picture quality}. Further, lifelong learning is used for unsupervised opinion target (aspect or entity) classification \cite{shu2016lifelong}, but not for aspect extraction. \cite{liu2016improving} proposed an unsupervised lifelong leaning method based on dependency rules~\cite{QiuLBC2011} and recommendation. However, it is different from our method as our method is based on supervised sequence labeling. We aim to find more precise aspects using supervised learning and show that lifelong learning is also effective for supervised learning and to propose a novel method to incorporate it into the CRF labeling process. 

There are existing lifelong supervised learning methods~\cite{chen2015lifelong,ruvolo2013ella} but they are for classification rather than for sequence labeling. 

Note that lifelong learning is related to transfer learning and multi-task learning~\cite{pan2010survey}, but they are also different. See their differences in~\cite{ChenLiu2014ICML}. 


\section{Conditional Random Field}
Conditional Random Field (CRF) is a popular supervised sequence labeling method. We use linear-chain CRF, which is the first order CRF. It can be viewed as a factor graph over an observation sequence $\mathbf{x}$ and a label sequence $\mathbf{y}$.

Let $L$ denote the length of the sequence, and $l$ indicate the $l$th position in the sequence. Let $\mathcal {F}=\{f_h(y_{l},y_{l-1},\mathbf{x}_l)\}_{h=1}^{H}$ be a set of feature functions. Each feature function $f_h$ has a corresponding weight $\theta_h$. The conditional probability of the sequence of labels $\mathbf{y}$ given the sequence of observations $\mathbf{x}=(\mathbf{x}_1, \dots, \mathbf{x}_L)$ is
\begin{equation}
\label{lccrf}
p(\mathbf{y}|\mathbf{x}; \boldsymbol{\theta}) = \frac{1}{Z(\mathbf{x})} \prod_{l=1}^{L} \exp \Bigg\{ \sum_{h=1}^{H}\theta_{h}f_{h}(y_{l},y_{l-1},\mathbf{x}_{l}) \Bigg\} ,
\end{equation}
where $Z(\mathbf{x})$ is the partition function:
\begin{equation}
\label{pf}
Z(\mathbf{x})=\sum_{\mathbf{y' \in \mathbf{Y}}}\prod_{l=1}^{L}\exp \Bigg\{ \sum_{h=1}^{H}\theta_{h}f_{h}(y_{l},y_{l-1},\mathbf{x}_{l}) \Bigg\} .
\end{equation}

\subsection{Parameter Estimation and Prediction}
During training, the weights $\boldsymbol{\theta} = \{\theta_1,
\dots, \theta_H\}$ of feature functions can be estimated by maximizing the log likelihood on the training data $D_t=\{(\mathbf{x}^{(1)},\mathbf{y}^{(1)}), \dots, (\mathbf{x}^{(n)},\mathbf{y}^{(n)})\}$: 
\begin{equation}
\label{learning}
\hat{\boldsymbol{\theta}} = \argmax_{\boldsymbol{\theta}} \log \prod_{i=1}^{n} p(\mathbf{y}^{(i)}|\mathbf{x}^{(i)}; \boldsymbol{\theta}) .
\end{equation}
During testing, the prediction of labels is done by maximizing
\begin{equation}
\label{predict}
\hat{\boldsymbol{\mathbf{y}}} = \argmax_{\boldsymbol{\mathbf{y}} \in \mathbf{Y}} p(\mathbf{y}|\mathbf{x}; \hat{\boldsymbol{\theta}}) .
\end{equation}

\subsection{Feature Function}
We use two types of feature functions. One is \emph{Label-Word (LW)} feature function:
\begin{equation}
\label{lw feature function}
f_{iv}^{\textit{LW}}(y_l,\mathbf{x}_l) = \mathbbm{1}\{y_l=i\}\mathbbm{1}\{\mathbf{x}_l=v\} ,\forall i \in \mathcal{Y}, 
\forall v \in \mathcal{V},
\end{equation}
where $\mathcal{Y}$ is the set of labels, $\mathcal{V}$ is the vocabulary and $\mathbbm{1}\{\cdot\} $ is indicator function. The above feature function returns $1$ when the $l$th word is $v$ and the $l$th label is $i$. Otherwise, it returns $0$. The other feature function is \emph{Label-Label (LL)} feature function:
\begin{equation}
\label{ll feature function}
f_{ij}^{\textit{LL}}(y_l,y_{l-1})=\mathbbm{1}\{y_l=i\}\mathbbm{1}\{y_{l-1}=j\} ,\forall i,j \in \mathcal{Y},
\end{equation}

Because the size of label set is small, the occurrence of each \textit{Label-Label} combination in the training data is sufficient to learn the corresponding weights for Eq. (\ref{ll feature function}) well. However, the set of observed words is much larger. The occurrence of each \textit{Label-Word} combination in the training data is not sufficient to ensure the corresponding weights are learned well for Eq. (\ref{lw feature function}). Further, the set of unobserved words is huge. When doing prediction (testing), it is highly possible that there are new words that have never appeared in the training data. So there are no corresponding \textit{Label-Word} feature function and weight that match those newly observed words. To solve this problem, we introduce additional features in the next section.

\section{Features}
In the \textit{Label-Word} feature function Eq (\ref{lw feature function}), $\mathbf{x}_l$ represents the current word that can take a value from a set of words. In practice, $\mathbf{x}_l$ is a multi-dimensional vector. We use $d \in \mathcal{D}$ to denote one feature (dimension) of $\mathbf{x}_l$, where $\mathcal{D}$ is the feature set of $\mathbf{x}_l$. The \emph{Label-dimension (L$d$)} feature function is defined as
\begin{equation}
\label{ld feature function}
f_{iv^d}^{\textit{L}d}(y_l,\mathbf{x}_l) = \mathbbm{1}\{y_l=i\}\mathbbm{1}\{\mathbf{x}_l^d=v^d\}, \forall i \in \mathcal{Y}, 
\forall v^d \in \mathcal{V}^d ,
\end{equation}
where $\mathcal{V}^d$ is the set of observed values in feature $d$ and we call $\mathcal{V}^d$ feature $d$'s feature values. Eq. (\ref{ld feature function}) is a feature function that returns $1$ when $\mathbf{x}_l$'s feature $d$ equals to the feature value $v^d$ and the variable $y_l$ ($l$th label) equals to the label value $i$.

For NLP problems, commonly used features are word (W) and POS-tag (P). The POS-tag feature is a more general feature than the word feature since it generalizes to new observations in testing. Contextual features in a fixed-sized window are useful as well, such as previous word (-1W), previous word's POS-tag (-1P), next word (+1W) and next word's POS-tag (+1P). 


The feature set we use for CRF is \{G, W, -1W, +1W, P, -1P, +1P\}, which contains 6 common features and 1 general dependency feature (G). The general dependency feature (G) takes a \emph{dependency pattern} as a value, which is generated from a dependency relation obtained from dependency parsing on input sentences. This is a useful feature because a dependency relation can link two words that may be far away from each other rather than a fixed-size window. The dependency feature also enables the capability of knowledge accumulation in lifelong learning, which will be clear shortly.  

\subsection{Dependency Relation}
A dependency relation is a 7-tuple of the following format: 
$$\textit{(type, gov, govidx, govpos, dep, depidx, deppos)}$$
where $\textit{type}$ is the type of the dependency relation, \textit{gov} is the \emph{governor word}, \textit{govidx} is the index (position) of the governor word in a sentence, \textit{govpos} is the POS tag of the governor word, \textit{dep} is the \emph{dependent word}, \textit{depidx} is the index of the dependent word in a sentence and \textit{deppos} is the POS tag of the dependent word.

\begin{table*}
	\centering
	\scalebox{0.80}{
	\begin{tabular}{ c | c| l }
		\hline
		Index & Word & Dependency Relations  \\
		\hline
		1 & The & \{\textit{(det, battery, 2, NN , The, 1, DT)} \}\\
		\hline
		2 & battery &\{\textit{(nsubj, great, 7, JJ , battery, 2, NN), (det, battery, 2, NN , The, 1, DT), (nmod, battery, 2, NN, camera, 5, NN)} \} \\
		\hline
		3 & of &\{\textit{(case, camera, 5, NN, of, 3, IN)} \}\\
		\hline
		4 & this&\{\textit{(det, camera, 5, NN, this, 4, DT)} \}\\
		\hline
		5 & camera&\{\textit{(case, camera, 5, NN, of, 3, IN), (det, camera, 5, NN, this, 4, DT), (nmod, battery, 2, NN, camera, 5, NN)} \} \\
		\hline
		6 & is&\{\textit{(cop, great, 7, JJ , is, 6, VBZ)} \}\\
		\hline
		7 & great&\{\textit{(root, ROOT, 0, VBZ, great, 7, JJ), (nsubj, great, 7, JJ , battery, 2, NN), (cop, great, 7, JJ , is, 6, VBZ)} \} \\
		\hline 
	\end{tabular}
	}
	\caption{Dependency relations parsed from ``The battery of this camera is great''}
	\label{table:dr1}
\end{table*}

Table \ref{table:dr1} shows the dependency relations parsed from ``The battery of this camera is great''. The Index column shows the position of each word in the sentence. The Dependency Relations column lists all the dependency relations that each word involves. 
 
The general dependency feature (G) of the variable $\mathbf{x}_l$ takes a set of feature values $\mathcal{V}^{\textit{G}}$. Each feature value $v^{\textit{G}}$ is a dependency pattern. The \emph{Label-G (LG)} feature function is defined as:
\begin{equation}
\label{lg feature function}
f_{iv^{\text{G}}}^{\textit{LG}}(y_l,\mathbf{x}_l) = \mathbbm{1}\{y_l=i\}\mathbbm{1}\{\mathbf{x}_l^{\textit{G}}=v^{\textit{G}}\} ,\forall i \in \mathcal{Y}, 
\forall v^{\text{G}}\in \mathcal{V}^{\text{G}}.
\end{equation}
Such a feature function returns $1$ when the general dependency feature of the variable $x_l$ equals to a dependency pattern $v^{\textit{G}}$ and the variable $y_l$ equals to the label value $i$.

\subsection{Generating Dependency Patterns}
As discussed above, the general dependency feature has possible values of all dependency relations. However, we do not use each dependency relation from the parser directly because it is too sparse to get better results during testing. We generalize a dependency relation into a {\em dependency pattern} using the following steps: 

\begin{enumerate}
	\item Eliminate all index information from a relation as the same word pattern may appear in different positions in different sentences. 
	After removing the index information, the dependency relation is in the following format:
	$$\textit{(type, gov, govpos, dep, deppos)} .$$
	\item Replace a specific word with a wildcard for all its dependency relations. There are two reasons for this. First, the word itself in a dependency relation is redundant since we already have word (W) and POS tag(P) features. Second, we care more about the other word's influence on the current word. We still keep the information whether a word is a dependent word or a governor word. This is because without such information, ``the battery of this camera'' has the same $\textit{nmod}$ relation as ``the camera of this battery''.
	
	To illustrate the process of this step, in Table \ref{table:dr1}, the 5th word ``camera'' is the dependent word in  $\textit{(nmod, battery, NN, camera, NN)}$ but the governor word in $\textit{(det, camera, NN, this, DT)}$. After applying wildcard, the relations for ``camera'' become:
	$$\textit{(nmod, battery, NN, *), (det, *, this, DT), (case, *, of, IN)} $$

	This is still not general enough because there are still actual words in the relations, which make the relations still too specific and difficult to apply to new domains (cross-domains) other than the training domain because those words may not appear in new domains. 

	\item Replace the related word in each dependency relation with a more general label to achieve a more general dependency feature value. Let the set of aspects annotated in the training data be $K_t$. If a word in the dependency relation appears in $K_t$, we replace it with a special label `A' (aspect) and if the word does not, it is replaced with the label `O' (other). 
	
	
	For example, assuming the training domain is \textit{Camera}, The words ``battery'' and ``camera'' are in $K_t$. The above dependency relations for the word ``camera'' become:
	$$\textit{(nmod, A, NN, *), (det, *, O, DT), (case, *, O, IN)} .$$
	Likewise, the dependency relations of the word ``battery'' become:
	$$\textit{(nsubj, O, JJ , *), (det, *, O, DT), (nmod, *, A, NN)} .$$
	
	These final forms of dependency relations are called {\em dependency patterns}. 
	
	In the sentence ``The battery of this camera is great'', the 5th word ``camera'' makes the feature function Eq. (\ref{lg feature function 1}) returns $1$ because ``camera'' is an aspect and it has a dependency pattern $\textit{(nmod, A, NN, *)}$ .	
	\begin{equation}
		\label{lg feature function 1}
		f_{iv^G}^{\textit{LG}}(y_l = \text{`A'},\mathbf{x}_l = \text{``camera''}) = 1, 
	\end{equation}
	where $i = \text{`A'}$ and $v^G = \textit{(nmod, A, NN, *)}$.
	Likewise, the 2nd word ``battery'' makes the feature function Eq. (\ref{lg feature function 2}) returns $1$ because it is an aspect as well and it has a dependency pattern $\textit{nmod(*, A, NN)}$.
	\begin{equation}
		\label{lg feature function 2}
		f_{iv^G}^{\textit{LG}}(y_l = \text{`A'}, \mathbf{x}_l = \text{``battery''}) = 1, 
	\end{equation}
	where $i = \text{`A'}$ and $v^G = \textit{(nmod, *, A, NN)}$.

\end{enumerate}

We are now ready to present the proposed Lifelong-CRF method since dependency patterns are capable of accumulating knowledge. 

	
	


\section{Lifelong CRF}

Due to the fact that dependency patterns for the general dependency feature do not use any actual words, they are powerful for cross-domain extraction (the test domain is not the training domain). More importantly, they make the proposed Lifelong-CRF method possible. The idea is as follows: 

We first introduce a set $K$ of {\em reliable aspects} which is mined from the aspects extracted from past domains datasets using a trained CRF model $M$. $K$ is regarded as the {\em past knowledge} in lifelong learning. $K$ is $K_t$ (the set of all annotated aspects in the training data $D_t$) initially. The more domains $M$ works on, the more aspects it extracts, and the larger the set $K$. When faced with a new domain, $K$ allows the general dependency feature to generate more dependency patterns related to aspects, which uses $K$ as we discussed in the previous section. More patterns generate more feature functions, which enable the CRF model $M$ to extract more and better aspects in the new domain.  

The proposed Lifelong-CRF algorithm works in two phases: training phase and lifelong prediction phase. In the training phrase, we train a CRF model $M$ using the annotated training data $D_t$. In the lifelong prediction phase, $M$ is applied to each new dataset for aspect extraction. In the lifelong process, $M$ does not change. Instead, as mentioned above, we keep a set $K$ of reliable aspects extracted from the past datasets. Clearly, we cannot use all extracted aspects from the past domains as reliable aspects due to many extraction errors. But those aspects that appear in multiple past domains are more likely to be correct. Thus $K$ contains those frequent aspects extracted in the past. Below, we discuss these two phases in greater detail. 

\textbf{Model Training Phase}: Given the annotated training data set $D_t$ and the set $K_t$ of all annotated aspects in $D_t$, it first generates all feature functions (including dependency pattern-based ones) to give the data $F$ with features (line \ref{train:1}). It then trains a CRF model $M$ by running a CRF learning algorithm (line \ref{train:2}). $K_t$ is assigned to $K$ as the initial set of reliable aspects (line \ref{train:3}), which will be used in subsequent extraction tasks in new domains. 
\begin{algorithm}
	\caption{Model Training Phrase}\label{algo:train}
	\begin{algorithmic}[1]
		\STATE $F \leftarrow \text{FeatureGeneration}(D_t, K_t)$ \label{train:1}
		\STATE $M \leftarrow \text{Train-CRF}(F)$ \label{train:2}
		\STATE $K \leftarrow K_t$	\label{train:3}
	\end{algorithmic}
\end{algorithm}

\textbf{Lifelong Prediction Phase}: This is the steady state phase for lifelong CRF prediction (or extraction). When a new domain dataset $D_{n+1}$ arrives in the system, it uses Algorithm \ref{algo:pred} to perform extraction on $D_{n+1}$, which works iteratively. 
\begin{enumerate}
	\item As in Algorithm \ref{algo:train}, it first generates the features on the data $D_{n+1}$ (line \ref{test:1}). It then applies the CRF model $M$ on $F$ to produce a set of aspects $A_{n+1}$ (line \ref{test:2}). It is important to note again that $K$ grows as the system worked on more domains, which enables the system to generate more dependency patterns-based feature functions for the new data, and consequently better extraction results from the new domain as we will see in the experiment section. 
	\item $A_{n+1}$ is added to $S$, our past aspect store. From $S$, we mine a set of frequent aspects $K_{n+1}$. The frequency threshold is $\lambda$.
	\item If $K_{n+1}$ is the same as $K_p$ from the previous iteration, the algorithm exits the loop as there will be no new aspects to be found. We now explain why we need an iterative process. This is because each extraction gives new results, which may increase the size of $K$, the reliable past aspects or the past knowledge. The increased $K$ may produce more dependency patterns, which may enable more extractions in the next iteration. 
	\item Else, this means that some additional reliable aspects are found. $M$ may be able to extract additional aspects in the next iteration. Lines \ref{test:3} and \ref{test:4} basically updates the two sets for the next iteration. Note that aspects $K_t$ from the training data are considered always reliable, thus a subset of $K$.
\end{enumerate}

\begin{algorithm}
	\caption{Lifelong Prediction Phase}\label{algo:pred}
	\begin{algorithmic}[1]
		\STATE $K_p \leftarrow \emptyset$
		\ALOOP{} 
		\STATE $F \leftarrow \text{FeatureGeneration}(D_{n+1}, K)$ \label{test:1}
		\STATE $A_{n+1} \leftarrow \text{Apply-CRF-Model}(M, F)$  \label{test:2}
		\STATE $S \leftarrow S \cup \{A_{n+1}\}$
		\STATE $K_{n+1} \leftarrow \text{Frequency}(S,\lambda)$
		\IF {$K_p = K_{n+1}$}
		\BREAK
		\ELSE
		\STATE $K \leftarrow K_t \cup K_{n+1}$\label{test:3}
		\STATE $K_p \leftarrow K_{n+1}$\label{test:4}
		\STATE $S \leftarrow S - \{A_{n+1}\}$
		\ENDIF
		\ENDALOOP
	\end{algorithmic}
\end{algorithm}

\section{Experiment}

{\bf Evaluation Datasets}: We use two types of data for our experiments. The first type consists of seven (7) annotated benchmark review datasets from 7 domains (products). Since they are annotated, they are used in training and testing. The first 4 datasets are from~\cite{HuL2004}, which actually has 5 datasets from 4 domains. Since we are mainly interested in results at the domain level, we did not use one of the domain-repeated datasets. The last 3 datasets of three domains (products) are from~\cite{liu2016improving}. All these datasets have been used previously for aspect extraction~\cite{HuL2004,Jakob2010,liu2016improving} Details of the datasets are in Table \ref{tab:annotation}.


The second type has 50 review datasets from 50 diverse domains or products~\cite{ChenLiu2014ICML}. These datasets are not annotated or labeled. They are used for lifelong learning and are treated as the past domain data. Note that since they are not annotated, we cannot use them for training of CRF or for testing. Each dataset has 1000 reviews. All these datasets were downloaded from the paper authors' webpages. 


\begin{table}[!]
	\centering
	\scalebox{0.95}{
	\begin{adjustbox}{max width=0.5 \textwidth}		
		\begin{tabular}{c|c|c|c|c}
			\hline
			{\bf Dataset}& {\bf Domain}&{\bf \# of Sentence}& {\bf \# of Aspect}& {\bf \# of Outside}  \\\hline
			D1 & Computer 	&536	&1173	&7675\\\hline
			D2 & Camera 	&609	&1640	&9849\\\hline
			D3 & Router 	&509	&1239	&7264\\\hline
			D4 & Phone 		&497	&980	&7478\\\hline
			D5 & Speaker  	&510	&1299	&7546\\\hline
			D6 & DVD Player &506	&928	&7552\\\hline
			D7 & Mp3 Player &505	&1180 	&7607\\\hline
		\end{tabular}
	\end{adjustbox}	
	}
	\caption{{\small Annotation details of the  benchmark datasets.}}\label{tab:annotation}
\end{table}

\begin{table*}[!t]
	\centering
	\caption{Comparative results on Aspect Extraction in precision, recall and F$_1$ score Cross Domain and In Domain ($-$X means all except domain X)}\label{tab:result}
	\begin{adjustbox}{max width=1.0\textwidth}	
		\begin{tabular}{c|c|c c c|c c c|c c c|c|c c c|c c c|c c c }
			\hline	
			&\multicolumn{10}{|c|}{\textbf{Cross Domain}}	&\multicolumn{10}{|c}{\textbf{In Domain}}\\\hline
			\multirow{2}{*}{Training}&\multirow{2}{*}{Testing}&\multicolumn{3}{ |c| }{CRF}& \multicolumn{3}{ |c| }{CRF+R}& \multicolumn{3}{ |c| }{Lifelong CRF} &
			\multirow{2}{*}{Testing}&\multicolumn{3}{ |c| }{CRF}& \multicolumn{3}{ |c| }{CRF+R}& \multicolumn{3}{ |c}{Lifelong CRF} \\
			&&$\mathcal{P}$& $\mathcal{R}$&$\mathcal{F}_1$&$\mathcal{P}$& $\mathcal{R}$&$\mathcal{F}_1$
			&$\mathcal{P}$& $\mathcal{R}$&$\mathcal{F}_1$
			&&$\mathcal{P}$& $\mathcal{R}$&$\mathcal{F}_1$&	$\mathcal{P}$& $\mathcal{R}$&$\mathcal{F}_1$ &	$\mathcal{P}$& $\mathcal{R}$&$\mathcal{F}_1$\\\hline
			$-$Computer&Computer&86.6&51.4&64.5&23.2&90.4&37.0&82.2&62.7&71.1&$-$Computer&84.0&71.4&77.2&23.2&93.9&37.3&81.6&75.8&78.6\\\hline
			$-$Camera&Camera&84.3&48.3&61.4&21.8&86.8&34.9&81.9&60.6&69.6&$-$Camera&83.7&70.3&76.4&20.8&93.7&34.1&80.7&75.4&77.9\\\hline
			$-$Router&Router&86.3&48.3&61.9&24.8&92.6&39.2&82.8&60.8&70.1&$-$Router&85.3&71.8&78.0&22.8&93.9&36.8&82.6&76.2&79.3\\\hline
			$-$Phone&Phone&72.5&50.6&59.6&20.8&81.2&33.1&70.1&59.5&64.4&$-$Phone&85.0&71.1&77.5&25.1&93.7&39.6&82.9&74.7&78.6\\\hline
			$-$Speaker&Speaker&87.3&60.6&71.6&22.4&91.2&35.9&84.5&71.5&77.4&$-$Speaker&83.8&70.3&76.5&20.1&94.3&33.2&80.1&75.8&77.9\\\hline
			$-$DVDplayer&DVDplayer&72.7&63.2&67.6&16.4&90.7&27.7&69.7&71.5&70.6&$-$DVDplayer&85.0&72.2&78.1&20.9&94.2&34.3&81.6&76.7&79.1\\\hline
			$-$Mp3player&Mp3player&87.5&49.4&63.2&20.6&91.9&33.7&84.1&60.7&70.5&$-$Mp3player&83.2&72.6&77.5&20.4&94.5&33.5&79.8&77.7&78.7\\\hline\hline
			&\textbf{Average}&82.5&53.1&64.3&21.4&89.3&34.5&79.3&63.9&70.5&\textbf{Average}&84.3&71.4&77.3&21.9&94.0&35.5&81.3&76.0&78.6\\\hline			
		\end{tabular}
	\end{adjustbox}	
\end{table*}


{\bf Compared Methods}: Since the goal of this paper is to study whether lifelong learning can be exploited to improve supervised learning for aspect extraction, we compare our proposed method Lifelong-CRF with the state-of-the-art supervised extraction methods. We will not compare with unsupervised extraction methods, which have been shown improvable by lifelong learning~\cite{liu2016improving}. This paper is the first to incorporate lifelong learning to supervised sequence labeling. Our experiment compares the following four methods.   

{\em CRF}: This is the linear chain CRF. We use the system from \footnote{https://github.com/huangzhengsjtu/pcrf/}. CRF has been used for the task by many researchers~\cite{Jakob2010,Choi2010,Mitchell-EtAl:2013:EMNLP}. 

{\em CRF+R}: This system treats the accumulated reliable aspect set $K$ in the past as a dictionary. It simply adds those reliable aspects in $K$ that are not extracted by CRF but are in the test data to the CRF results. We want to see whether incorporating $K$ into the CRF model for prediction in Lifelong-CRF is actually needed.     


{\em Lifelong-CRF}: This is our proposed system. The frequency threshold $\lambda$ in Algorithm 2 used in our experiment to judge which extracted aspects are considered reliable is empirically set to $2$. 

{\bf Experiment Setting}: In order to compare the systems using the same training and test data, we split each dataset into three parts, 200 sentences for training, 200 sentences for testing. 

In our experiments, we conducted both cross-domain and in-domain tests. We are particularly interested in cross-domain tests as it is labor-intensive and time-consuming to label training data for each domain. It is thus highly desirable to have the trained model used in cross-domain situations. Also, training and testing on the same domains (in-domain) is less interesting because those aspects appeared in the training data are very likely to appear in the test data because they are all about reviews of the same products. 

{\em Cross-domain experiments}: We combine 6 datasets for training (1200 sentences), and then test on the 7th domain (not used in training). This gives 7 {\em cross-domain} results. 

{\em In-domain experiments}: We train and test on the same 6 domains (1200 sentences for training and 1200 sentences for testing). This gives us 7 {\em in-domain} results as well. Note that although we call these in-domain experiments, both the training and testing data use the same 6 domains. 

{\bf Evaluating Measures}: 
Since our goal is to extract aspects, we use the popular precision $\mathcal{P}$, recall $\mathcal{R}$, and $\mathcal{F}_1$-score to evaluate our results on the extracted aspects.  

\subsection{Results Analysis}

All the experiment results are given in Table \ref{tab:result}. We analyze the results of cross-domain and in-domain in turn. Cross-domain is the main setting of interest as it is very undesirable to manually label data for every domain in practice. 

\begin{enumerate}
	\item {\em Cross-domain results analysis}. Each entry $-$X in the first column means that domain X data is not used in training, i.e., the other 6 domains are used in training for the experiment. For example, $-$Computer means that the data from the Computer domain is not used in training. In the second column, X means that domain X is used in testing. 
	
	From the cross-domain results in the table, we observe the following: In $\mathcal{F}_1$ score, Lifelong-CRF is much better than CRF. On average the $\mathcal{F}_1$ score improves from $64.3$ to $70.5$. This is a major improvement. The main improvement is on the recall, which are markedly better. 
CRF+R's results are very poor due to poor precisions, which show that treating the reliable aspects set $K$ as a dictionary is a bad idea. Incorporating $K$ into the CRF model is important because many aspects in $K$ are not correct or not applicable to the new/test domain. CRF model $M$ will not extract many of them. 
	
	\item {\em In-domain results analysis}: Training uses the same 6 domain data as in the cross-domain case. $-$X in the Testing column of the in-domain results means that X is not used in testing. For example, in the first row under in-domain, the Computer domain is not used in training or testing. That is, the other 6 domains are used in both training and testing (thus in-domain). 
	
	From the in-domain results in the table, we observe the following: Lifelong-CRF still improves CRF, but the amount of improvement is considerably smaller. This is expected as we discuss above, because most of the aspects appeared in training probably also appeared in the test data because they are reviews from the same 6 products. Again CRF+R does poorly due to the same reason. 
	
\end{enumerate}

\section{Conclusion}

This paper proposed a lifelong learning based approach to enabling Conditional Random Fields (CRF) to leverage the past knowledge gained from extraction results of multiple domains to improve CRF's extraction performance. To our knowledge, this is the first time that CRF is incorporated with the lifelong learning capability. This is also the first time that a lifelong supervised method is used for aspect extraction in opinion mining. Experimental results demonstrated the superior performance of the proposed Lifelong-CRF method. 


\bibliography{emnlp2016}
\bibliographystyle{emnlp2016}

\end{document}